\documentclass[]{article}
\usepackage{proceed2e}

\usepackage{times}

\usepackage{algorithmic}
\usepackage{algorithm}
\usepackage{amsmath}
\usepackage{amsthm}
\usepackage{amssymb}
\usepackage{kbordermatrix}
\usepackage{hyperref}       
\usepackage{url}            
\usepackage{booktabs}       
\usepackage{amsfonts}       
\usepackage{nicefrac}       
\usepackage{microtype}      
\usepackage{graphicx}
\usepackage{natbib}

\graphicspath{{./figures/}}

\DeclareMathOperator{\FitPoly}{FitPolynomial}
\DeclareMathOperator{\mean}{mean}
\DeclareMathOperator{\threshold}{threshold}

\newtheorem{definition}{Definition}

\title{Dynamical Kinds and their Discovery}

\author{} 

\author{ {\bf Benjamin C. Jantzen} \\
Philosophy Dept. \\
Virginia Tech\\
Blacksburg, VA 24061 
}

\begin{document}

\maketitle

\begin{abstract}
We demonstrate the possibility of classifying causal systems into kinds that share a common structure without first constructing an explicit dynamical model or using prior knowledge of the system dynamics. The algorithmic ability to determine whether arbitrary systems are governed by causal relations of the same form offers significant practical applications in the development and validation of dynamical models. It is also of theoretical interest as an essential stage in the scientific inference of laws from empirical data. The algorithm presented is based on the dynamical symmetry approach to dynamical kinds. A dynamical symmetry with respect to time is an intervention on one or more variables of a system that commutes with the time evolution of the system. A dynamical kind is a class of systems sharing a set of dynamical symmetries. The algorithm presented classifies deterministic, time-dependent causal systems by directly comparing their exhibited symmetries. Using simulated, noisy data from a variety of nonlinear systems, we show that this algorithm correctly sorts systems into dynamical kinds. It is robust under significant sampling error, is immune to violations of normality in sampling error, and fails gracefully with increasing dynamical similarity. The algorithm we demonstrate is the first to address this aspect of automated scientific discovery.
\end{abstract}

\section{The problem}
Recent decades have seen remarkable progress in addressing the problem of learning causal structure from observational or experimental data. One especially rich approach deploys the framework of graphical causal models (GCMs) \citep{pearl_causality_2000, spirtes_causation_2000, spirtes_causal_2016}. Given a particular set of variables describing a specific system or population, the learning problem is, roughly, to ascertain the set of direct, possibly stochastic causal relations amongst variables and, often, a model of the explicit functional form of those causal dependencies. Similarly, there is vast literature on system identification that focuses on producing a model of a given complex dynamical system that is suitable for prediction and control \citep{norton_introduction_1986, ljung_perspectives_2010}. While these lines of research differ in terminology, methods, and emphasis -- system identification tends to focus on deterministic systems, while the GCM formalism is designed for representing stochastic causal relations -- they both provide methods for discovering causal structure. They also share an emphasis on the individual system -- particular systems of causal relations are grouped together into categories only after the fact, and generally without a consistent aim or motivation.

This exclusive focus on individual systems, however, leaves out a significant element of scientific inference. Much scientific work, especially in the early exploration of a new phenomena, involves grouping distinct physical systems into categories. These categories are then targets for law-like generalization. In other words, a primary goal of scientific research is the identification of laws of \emph{kinds} of system, not just models of one-off systems. So, for example, we seek laws of chemical kinetics, not just models of the particular reaction in some specific beaker on a lab bench. We seek laws of motion, not just a model of how this particular pendulum moves given a variable arm length. This sort of inference to kinds is not reflected in the causal learning or system identification algorithms that have been developed -- both approaches to learning causal structure allow for classification only after model creation.

But what if such classification could be done prior to learning detailed models of the causal structure of individual systems? In that case, a number of inferential tasks would become substantially easier. For example, if one knew in advance that multiple systems are governed by a common type of causal structure --  whether representable by, say, a parametric class of differential equations or an equivalence class of GCMs -- then one could pool data from multiple systems in order to determine the form of the differential equation or an equivalence class of GCMs governing systems of that sort. Or consider the problem of  model validation. For sufficiently complex computer models, validation can be exceedingly difficult because the dynamical relations amongst high-level variables in the simulation are unknown \citep{mattingly_projectible_2009}. But if it were possible to ascertain directly -- without an explicit model of the high-level dynamics in either the simulation or the target system -- that the dynamics are of the same type, this would provide a tractable route to validation. Finally, one could assess whether or not a system with unknown dynamics exhibits transitions between different dynamical regimes, such as the transition from laminar to turbulent flow in fluid dynamics. In such transitions, one and the same system exhibits a different dynamical form or causal structure at different times or in different circumstances. 

For these reasons, a method for assessing the sameness or difference of dynamical structure in the absence of any explicit dynamical model would be both practically and theoretically valuable. For such a direct classification to be possible, two things are required. First, a clear and rigorous definition of dynamical kind, and second, an empirical test for sameness of kind. We suggest that the first component -- a rigorous definition -- has been provided and well-motivated by the theoretical work of \citet{jantzen_projection_2014}. To meet the second need, we present here both a test derived from this notion of dynamical kind and an explicit algorithm for automating that test that is accurate and robust in the presence of significant measurement error in empirical data.

%
%
%
%
%
%
%
%
%
%
%
%
%

\subsection{Directly discerning dynamical difference: an example}

To get a sense for how such a test might work, consider two different species of bacteria, A and B, in dilute solutions of nutrient-rich media. One might be interested in knowing whether the growth of both species is governed by the same dynamics under these conditions, as suggested by the ``First Law of Ecology'' \citep{turchin_does_2001} . Specifically, one might be interested in whether both systems are well-represented by differential equations of the same form, whatever that form may be. The question can be approached directly by considering the sorts of properties shared by two systems governed by equations of the same form. One such property is the behavior of the system under intervention. More specifically, for any given dynamics there is a special class of transformations of the system state that commute with the evolution of that system state through time. 

To be concrete, suppose species A grows exponentially, such that $dx_A(t)/dt = r_A x_A(t)$. In that case, all and only scaling transformations of the form $\sigma_k(x_A(t)) = k x_A(t)$ commute with the time-evolution of the system so that it doesn't matter whether one applies $\sigma_k$ and then evolves the system or evolves the system and then applies $\sigma_k$. Note that this is equivalent to the condition that $d\sigma_k(x_A(t))/dt = r_A \sigma_k(x_A(t))$. Even if these transformations were unknown to us, there is a simple way to ascertain whether the $\sigma_k$ for species A are different from the corresponding transformations, $\tau$, that commute with the time evolution of species B. To do so, we can read these transformations -- at least over a finite interval -- directly off of our experimental data. Suppose we make simultaneous measurements of each population at $m$ times, $t_0, t_1, \dots, t_m$. Suppose further that, for each species, two experiments are conducted. In the first, an intervention sets the initial populations to $x_A(t_0)$ and $x_B(t_0)$ for species A and B, respectively, and in the second we set the initial populations to $\tilde{x}_A(t_0)$ and $\tilde{x}_B(t_0)$. For each population, the resulting pair of trajectories must be connected by one of the special time-commuting transformations characteristic of that system. Empirical curves for $\sigma$ and $\tau$ can thus be constructed by plotting $\tilde{x}_A(t_i)$ as a function of $x_A(t_i)$ and $\tilde{x}_B(t_i)$ as a function of $x_B(t_i)$. To decide whether the dynamics of the two systems have the same form it is only necessary to determine whether $\sigma(x_A)\equiv \tilde{x}_A(x_A) = \tilde{x}_B(x_B) \equiv \tau(x_B)$. If species A really grows exponentially, then $\tilde{x}_A(x_A)=kx_A$, and detecting a difference in dynamics depends only on detecting deviation from a straight line in the sample of $\tilde{x}_B(x_B)$.

\section{Theoretical background and related work}

\subsection{Theory of dynamical kinds}

The preceding example appealed to a feature of the dynamics of all exponential growth systems. That is, dynamical systems were grouped into categories on the basis of high-level features of their dynamics -- in this case, the form of the governing differential equation -- rather than features of specific trajectories or solutions of that equation. This is the essence of a \emph{dynamical kind} -- a class of causal systems that share some salient higher-order features though they may differ dramatically in the specific relations amongst variables. 

Of course, there are indefinitely many higher-order features of causal structure that could be used for classification in this way. We have chosen to focus on the variety of higher-order properties that were used to sort dynamical kinds in the bacterial growth example. Specifically, the notion of dynamical kind we propose to exploit for purposes of automated discovery is that proposed by \citet{jantzen_projection_2014}. The motivation is two-fold. First, Jantzen's proposal picks out classes of causal system that, if identified in the absence of detailed individual models, would provide the sort of practical benefits for data-pooling, validation, and regime-detection described above. Second, as Jantzen argues in detail, the classes carved out in this way coincide well with existing scientific practice. Thus, automating the discovery of dynamical kinds in Jantzen's sense would go a long way toward filling a gap in automated scientific inference, namely the absence of algorithmic methods for determining classes of system appropriate for framing laws.

In Jantzen's view, what makes the dynamics of two distinct systems members of the same kind is a higher-order property that they share: the structured set of transformations of each variable that commute with incrementation of another variable. These transformations are called `dynamical symmetries' and are defined as follows \citep[p3633]{jantzen_projection_2014}:

\begin{definition}[Dynamical symmetry]
	Let $V$ be a set of variables. Let $\sigma$ be an intervention on the variables in $Int \subset V$. The transformation $\sigma$ is a dynamical symmetry with respect to some index variable $X \in V-Int$ if and only if $\sigma$ has the following property: for all $x_i$ and $x_f$, the final state of the system is the same whether $\sigma$ is applied when $X = x_i$ and then an intervention on $X$ makes it such that $X = x_f$, or the intervention on $X$ is applied first, changing its value from $x_i$ to $x_f$, and then $\sigma$ is applied.
\end{definition}

Notice that this definition is substantive rather than mathematical in that it appeals to physical interactions of a certain sort, namely interventions in the GCM sense. As such, this definition of dynamical symmetry is not tied to any particular formalism or application. The notion of commuting interventions is certainly at home in the GCM framework, but it is also easily translated into the language of differential equations as the commutation of system evolution and the manipulation of initial or boundary conditions. It is the latter framework we focus on in the remainder of this paper. In particular, we will be concerned with classes of system that are well-described by differential equations. In fact, we'll further restrict attention to systems with explicit time dependence. In this special case of temporal dynamics, time is a uniquely natural ``index variable'' -- in a sense, one can't help but ``intervene'' to increment time in any given experiment. Thus, we are primarily interested in interventions on subsets of variables which are dynamical symmetries with respect to time \citep[p3632]{jantzen_projection_2014}:

\begin{definition}[Dynamical symmetry with respect to time]
	\label{def:DST}
	Let $t$ be the variable representing time, and let $V$ be a set of additional dynamical variables such that $t \notin V$. Let $\sigma$ be an intervention on the variables in $Int \subset V $. The transformation $\sigma$ is a dynamical symmetry with respect to time if and only if for all intervals $\Delta t$, the final state of the system is the same whether $\sigma$ is applied at some time $t_0$ and the system evolved until $t_0 + \Delta t$, or the system first allowed to evolve from $t_0$ to $t_0 + \Delta t$ and then $\sigma$ is applied.
\end{definition}

To put this in terms familiar from mathematical physics \citep{rosen_symmetry_1995, rosen_symmetry_2008}, let $\Lambda_{t,s}(\mathbf{x})$ be the propagator for a given system, i.e., the function $\Lambda_{t,s}: X \rightarrow X$ for which $\Lambda_{t,s}(\mathbf{x})$ is the state of the system at time $t$ given that $\mathbf{x} \in X$ is the state at time $s$. A function $\sigma: X \rightarrow X$ represents a dynamical symmetry with respect to time if and only if for every $s, t>s, \text{ and } x$, $\Lambda_{t,s}(\sigma(\mathbf{x})) = \sigma(\Lambda_{t,s}(\mathbf{x}))$. Put this way, it's clear that most symmetries of interest to physicists are dynamical symmetries with respect to time. Assuming that the dynamics is deterministic and well-described by a system of ordinary differential equations in time, we can also restate the condition in terms of flows \citep{logemann_ordinary_2014}. These are mappings $\phi: X \times \mathbf{R} \rightarrow X$ where $X$ is the space of sates of the system at times represented by the reals and under the conditions that $\phi(\mathbf{x},0)=0$ and $\phi(\phi(\mathbf{x},t),s)=\phi(\mathbf{x}, t+s)$. Dynamical symmetries in the framework correspond to functions $\sigma$ such that, for all $\mathbf{x}$ and $t$:

\begin{equation}
	\label{eq:flow}
	\phi(\sigma(\mathbf{x}),t) = \sigma(\phi(\mathbf{x},t))
\end{equation}

Whatever formalism is appropriate for expressing dynamical symmetries in a particular application, it is not necessarily true that all functions satisfying a given formal condition are proper dynamical symmetries. The latter are defined in terms of interventions on variables -- interactions with a system that change its measured quantities in a way that is at least physically possible and which does not destroy the causal connections amongst them. Some formal transformations may represent unrealizable interventions such as, perhaps, time reversal. For deterministic dynamics represented by a system of ODEs in time, every dynamical symmetry can be represented by a mapping satisfying the condition expressed by Equation \ref{eq:flow}. But if we take Jantzen's definition seriously, then it may not be the case that every such $\sigma$ represents a dynamical symmetry. However, since it is impossible to give a general account of what these exceptions are, we will make the idealizing assumption that, for the varieties of system under analysis here, there is a one-to-one correspondence between dynamical symmetries and functions satisfying Equation \ref{eq:flow}.

In Jantzen's sense, a bare set of dynamical symmetries is insufficient to characterize a dynamical kind. Further distinctions can be made if we note that for any given system of causally connected variables, the corresponding set of dynamical symmetries exhibits structure under composition. In other words, one dynamical symmetry applied after another is also a dynamical symmetry, and the equivalence relations amongst such compositions constitute a symmetry structure \citep[p3634-3635]{jantzen_projection_2014}:

\begin{definition}[Nontrivial symmetry structure]
	The \emph{symmetry structure} of a collection of dynamical symmetries, $\Sigma=\{\sigma_i|i=1,2,\dots\}$ is given by the composition function $\circ : \Sigma \times \Sigma \rightarrow \Sigma$. A \emph{nontrivial symmetry structure} is one that contains the identity (i.e., \lq\lq{}do nothing\rq\rq{}) transformation and, for at least one variable $Y$ relative to some index variable $X$, is \emph{not} isomorphic to the group of mappings from $Y$ to itself (i.e., the set of all transformations of $Y$).
\end{definition}

Jantzen identifies these nontrivial symmetry structures with the dynamical kinds of interest \citeyearpar[p3635]{jantzen_projection_2014}:

\begin{definition}[Dynamical kind]
A \emph{dynamical kind} is a class of systems of variables that share a set of dynamical symmetries that are related by a non-trivial symmetry structure. 
\end{definition}

\subsection{Related work}
As suggested above, Jantzen's theory of dynamical kinds is closely connected to the axiomatic treatment of causation that underwrites the increasingly diverse and powerful set of causal discovery algorithms. In particular, Jantzen shows that a system of variables possesses a nontrivial symmetry structure $S$ (one that is \emph{not} isomorphic to the set of automorphisms on the states of $S$) if and only if it has a non-trivial causal structure (one for which at least one variable is a direct cause of another)\citeyearpar[p3635-3636]{jantzen_projection_2014}. This is the sense in which dynamical symmetries are high-level properties of the causal structure of a system. There is also precedent for Jantzen's appeal to symmetries in the machine learning literature. \citet{schmidt_distilling_2009} developed an automated discovery algorithm that learns free-form laws by seeking invariants of the motion -- functions of the dynamical variables whose values remain constant through time. As Noether \citep{noether_invariant_1971} demonstrated, the presence of symmetries is intimately connected with the presence of such invariants. But to the best of our knowledge, there have been no prior attempts to use empirical measurements of dynamical symmetries to sort systems in the absence of detailed causal models.

\subsection{Learning dynamical kinds}
Though it is not sufficient, sharing a set of dynamical symmetries is a necessary condition for two systems to belong to the same dynamical kind. Given that the collection of possible dynamical symmetries is uncountably large as is the set of interventions one could perform on the system, verifying that any two systems have all of their dynamical symmetries in common is not possible. However, it is possible to verify with confidence that two systems do $not$ share all of their symmetries, and thus differ with respect to dynamical kind (and ultimately causal structure). The algorithm described in the next section uses empirical data from systems that are presumed to be governed by a deterministic temporal dynamics in order to assess whether or not the systems being compared differ with respect to one or more dynamical symmetry with respect to time. 

\begin{figure}
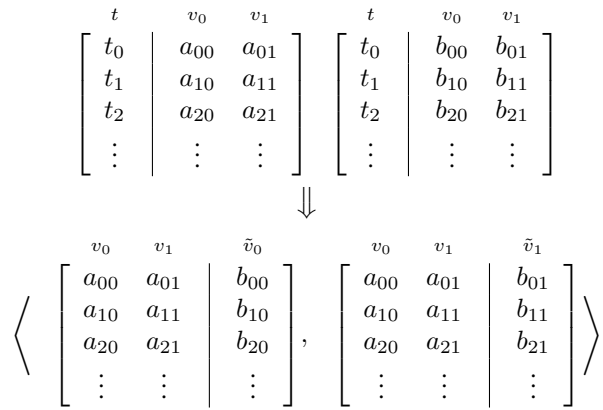

	\label{fig:data}
	\begin{center}
\begin{gather*}
	\kbordermatrix{ & t &  & v_0 & v_1\\
					& t_0 & \vrule & a_{00} & a_{01}\\
					& t_1 & \vrule & a_{10} & a_{11}\\
					& t_2 & \vrule & a_{20} & a_{21}\\
					& \vdots & \vrule & \vdots & \vdots\\
	}  
	\kbordermatrix{ & t &  & v_0 & v_1\\
					& t_0 & \vrule & b_{00} & b_{01}\\
					& t_1 & \vrule & b_{10} & b_{11}\\
					& t_2 & \vrule & b_{20} & b_{21}\\
					& \vdots & \vrule & \vdots & \vdots\\
	}\\
	\big \Downarrow\\
	\Biggl \langle
	\kbordermatrix{ & v_0 & v_1 & & \tilde{v}_0\\
					& a_{00} & a_{01} & \vrule & b_{00}\\
					& a_{10} & a_{11} & \vrule & b_{10}\\
					& a_{20} & a_{21} & \vrule & b_{20}\\
					& \vdots & \vdots & \vrule & \vdots\\
	},
	\kbordermatrix{ & v_0 & v_1 & & \tilde{v}_1\\
					& a_{00} & a_{01} & \vrule & b_{01}\\
					& a_{10} & a_{11} & \vrule & b_{11}\\
					& a_{20} & a_{21} & \vrule & b_{21}\\
					& \vdots & \vdots & \vrule & \vdots\\
	}\Biggr \rangle
\end{gather*}
\caption{Schematic showing how samples from two time-series are restructured to obtain an implicit model of the dynamical symmetry that maps one trajectory into the other.}
\end{center}
\end{figure}

\section{The algorithm}
\label{sec:algo}
Comparing dynamical systems proceeds in three phases: (i) sampling, (ii) transformation, and (iii) comparison. In phase (i), the variables $v_1, \dots, v_n$ are measured at times $t_1, \dots, t_m$ for both System 1 and System 2, starting from $p \geq 2$ distinct initial conditions (the set of initial conditions is presumed to be identical for the two systems). The result, for each system, is a sample of $p$ vector functions, each with $n$ components. In the second phase, these samples are transformed in order to serve as samples of the symmetry transformations to be compared. These dynamical symmetries correspond to the $p-1$ vector functions, $\tilde{\vec{x}}_j(\vec{x_1}(t))$ which map the initial curve $\vec{x}_1(t)$ into each of the other sampled curves (so $j \in {2,\dots,p})$. Any given symmetry function can thus be decomposed into $n$ functions of $n$ variables: $\langle \tilde{x}_{j,1}(x_{1,1},\dots,x_{1,n}),\dots,\tilde{x}_{j,n}(x_{j,1},\dots,x_{j,n})\rangle$ . To estimate these components for each symmetry, the sampled data is rearranged as follows. Using time to match up corresponding samples, the values of $\langle x_{1,1}(t), \dots, x_{1,n}(t) \rangle$ are paired separately with each component of $\tilde{\vec{x_i}}(t)$. This is shown schematically for $n = p = 2$ in Figure 1. The result is a set of $n$ matrices for each of the $p-1$ dynamical symmetries implicitly measured in the experiments.

In the final phase, the sets of transformed data representing specific dynamical symmetries are compared across systems. The output is a judgment that one or more dynamical symmetries differ between System 1 and System 2 (the positive case), or that the set of symmetries is indistinguishable (the negative case). To make this judgment, 10-fold cross-validation is used to estimate the error of two competing models, \emph{sep} and \emph{joint}. For the \emph{sep} model, data from System 1 and System 2 are treated separately. Each of the $n$ components of the $p-1$ symmetry transformations for System 1 are fit with polynomial surfaces independently of the $n$ components of the $p-1$ symmetries of System 2. For the \emph{joint} model, the data from System 1 and System 2 for each of the $p-1$ symmetries is combined and treated as a single sample from the same function. 

For both models, polynomial surfaces are used to provide a global fit of each component function, $\tilde{\vec{x}}_{j,i}(\vec{x}_1(t))$ . These surfaces are fit using singular-value decomposition to find the least-squares parameter values for a polynomial in $n$ variables and order $d$ \citep{golub_singular_1970}. The order of the surface is determined by 10-fold cross-validation. Specifically, the error for a fit of order 1 is estimated via 10-fold cross-validation. Higher orders continue to be investigated until the relative reduction in estimated error fails to exceed a tunable threshold, $\epsilon$, at which point the previous order considered is used to fit the entire dataset. For the results reported here, $\epsilon$ was fixed at $10^{-4}$, though as Figure \ref{fig:NS} (e)-(f) indicates, the performance of the algorithm is largely insensitive to the value of epsilon over ten orders of magnitude. 

The estimated error of \emph{sep} and \emph{joint} cannot be directly compared. If in fact the two systems are of the same dynamical type and thus have all of their symmetries in common, then the two models are expected to have the same error. However, due to random noise in the data and random variation in how the data is partitioned, cross-validation will provide slightly different estimates. To meaningfully compare the competing models, it must be determined how much of the difference in error is attributable to noise alone. In order to do so, the data from System 1 is divided in two, and the same 10-fold cross-validation routine is used to estimate the errors of joint and combined models. In this case, since all of the data comes from the same system, any difference in estimated error is due to noise in the data. Call the absolute value of this difference, $\delta_1$. The procedure is repeated with the data from System 2 to find $\delta_2$. Then $\delta = \max(\delta_1, \delta_2)$ is used as a threshold to compare \emph{sep} and \emph{joint}. Specifically, if the mean squared error of the \emph{joint} model as estimated by cross-validation exceeds the mean square error estimate for the \emph{sep} model by more than $\delta$, i.e. $MSE_{joint} > MSE_{sep} + \delta$, then the hypotheses of shared symmetries is rejected and the systems are judged to be of different kinds.  The overall algorithm for the comparison phase is shown in Algorithm \ref{table:outerCV}.

\begin{algorithm}
\caption{Dynamical Type Comparison}
\label{table:outerCV}
\begin{algorithmic}[1]
	\REQUIRE $A_{\sigma, v}$ -- for each transformation, $\sigma$ and target variable, $v_i$, measured for \textbf{System 1}, contains an $m \times (n+1)$ matrix with $m$ samples (rows) of $v_1, \dots, v_n$ and the value of $v_i$ to which $\sigma$ maps the system state\\
		$B_{\sigma, v}$ -- for each transformation, $\sigma$ and target variable, $v$, measured for \textbf{System 2}, contains an $m \times (n+1)$ matrix with $m$ samples (rows) of $v_1, \dots, v_n$ and the value of $v$ to which $\sigma$ maps the system state
	\ENSURE a boolean value indicating whether the data belong to systems of different dynamical types
	\STATE for each $\sigma$ and $v$, randomize the rows of $A_{\sigma, v}, B_{\sigma,v}$
	\STATE $Partition^1_{\sigma, v} \gets$ divide the rows of $A_{\sigma, v}$ into 10 segments
	\STATE $Partition^2_{\sigma, v} \gets$ divide the rows of $B_{\sigma, v}$ into 10 segments
	\FOR{$i = 1$ \TO 10}
	\STATE $Train^1_{\sigma, v} \gets \bigcup \lbrace Partition^1_{\sigma,v}[j] | j \neq i \rbrace$
	\STATE $Train^2_{\sigma, v} \gets \bigcup \lbrace Partition^2_{\sigma,v}[j] | j \neq i \rbrace$
	\STATE $TrainJoint_{\sigma, v} \gets Train^1_{\sigma, v} \cup Train^2_{\sigma, v}$
	\STATE $Model^1_{\sigma, v} \gets \FitPoly(Train^1_{\sigma, v})$
	\STATE $Model^2_{\sigma, v} \gets \FitPoly(Train^2_{\sigma, v})$
	\STATE $Model^{1,2}_{\sigma, v} \gets \FitPoly(Train^1_{\sigma, v} \cup Train^2_{\sigma, v})$
	\STATE $SE^1 \gets \bigcup \lbrace$for all $\sigma, v$, the squared errors of $Model^1_{\sigma, v}$ with respect to $Partition^1_{\sigma, v}[i]\rbrace$ 
	\STATE $SE^2 \gets \bigcup \lbrace$for all $\sigma, v$, the squared errors of $Model^2_{\sigma, v}$ with respect to $Partition^2_{\sigma, v}[i]\rbrace$ 
	\STATE $SE^{1,2} \gets \bigcup \lbrace$for all $\sigma, v$, the squared errors of $Model^{1,2}_{\sigma, v}$ with respect to $Partition^1_{\sigma, v}[i] \cup Partition^2_{\sigma, v}[i]\rbrace$ 
	\ENDFOR
	\STATE $MSEsep \gets \mean(SE^1 \cup SE^2)$
	\STATE $MSEjoint \gets \mean(SE^{1,2})$
	\IF{$MSEjoint > MSEsep + \threshold(A_{\sigma, v}, B_{\sigma, v})$}
	\STATE return $\mathbb{TRUE}$
	\ELSE
	\STATE return $\mathbb{FALSE}$
	\ENDIF
\end{algorithmic}
\end{algorithm}

\begin{figure}[!ht]
	\label{fig:LG}
	\centering
	\includegraphics[width=\columnwidth]{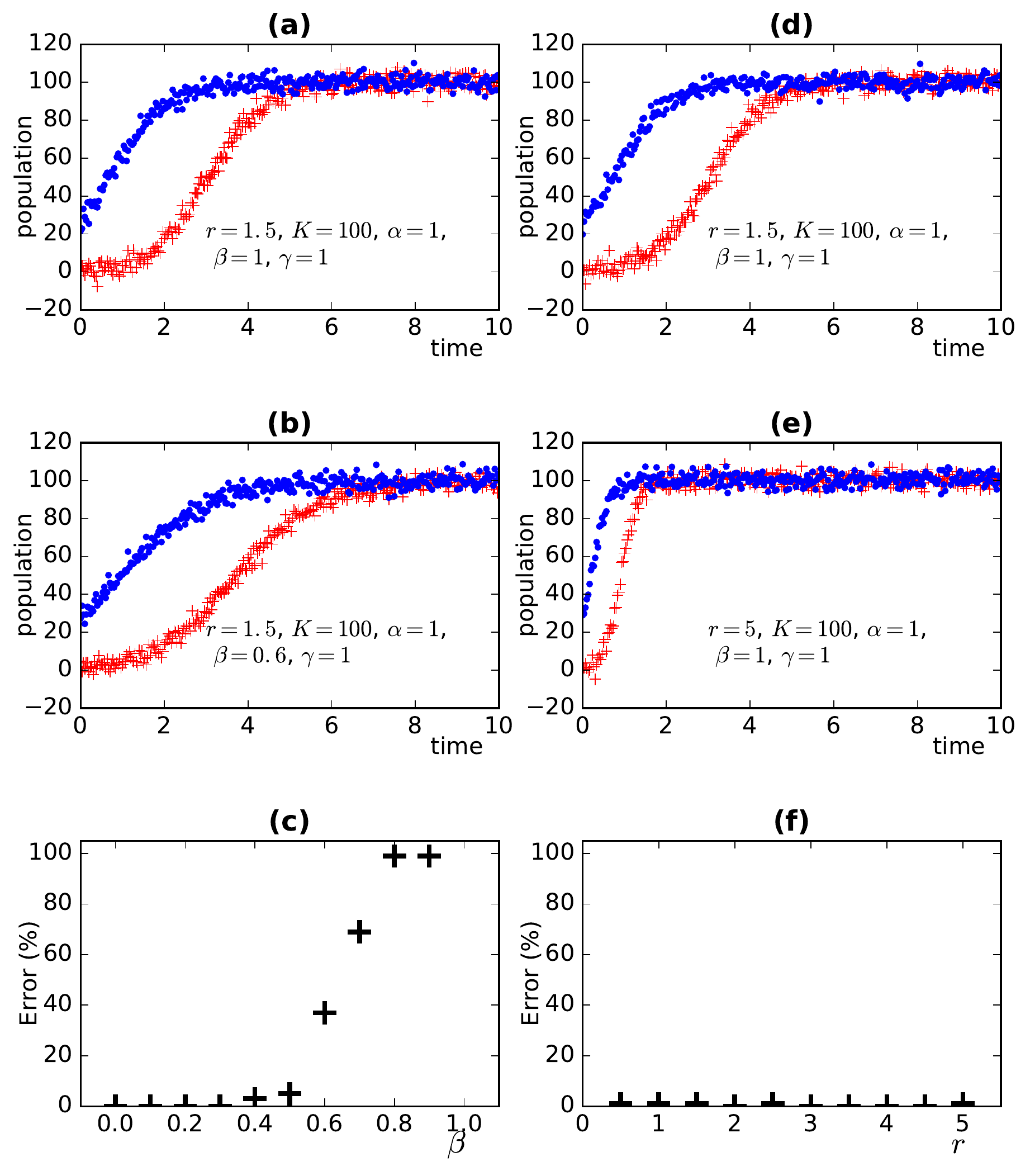}
	\caption{(a) A representative sample of data from the simulated logistic growth system with the indicated parameter values and Gaussian noise with a standard deviation of 3. The two curves correspond to initial populations of 1 (red pluses) and 25 (blue dots). (b) Data sampled from a growth system belonging to a \emph{different} dynamical kind. (c) The error rate of the algorithm as the value of $\beta$ is varied in one of the models compared, while fixed at $\beta = 1$ for the other. (d) and (e) Representative samples of growth systems of the \emph{same} dynamical kind. (f) The error rate of the algorithm as a function of the value of $r$ in the underlying generalized logistic growth model, with one system held fixed at $r=1.5$.}
\end{figure}

\begin{figure}[!ht]
	\label{fig:L-V}
	\includegraphics[width=\columnwidth]{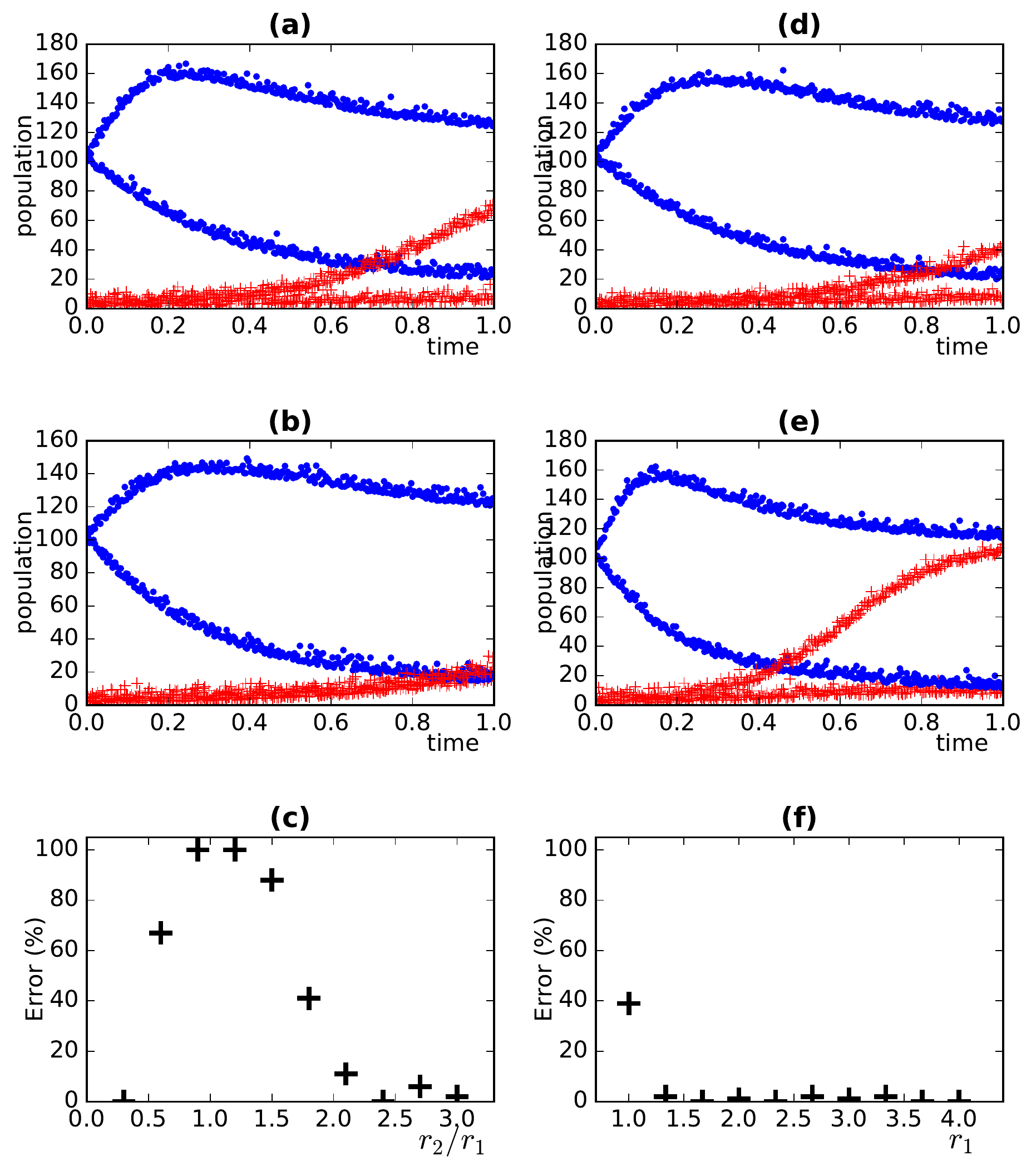}
	\caption{(a) Representative sample of data from the simulated two-species competitive Lotka-Volterra model with noise drawn from a Gaussian distribution with standard deviation of 5. The trajectories marked with pluses correspond to both competing species with initial populations of 1, and the trajectories marked with dots indicate the growth of both species starting from initial populations of 25 each. (b) Data sampled from a system belonging to a \emph{different} kind, (c) The error rate of the model as the ratio of growth rates $r_2/r_1$ is varied in one of the systems compared, while the other system is held fixed at $r_1 = r_2 = 2$. (d), (e) Representative samples from a pair of systems of the \emph{same} dynamical kind but with distinct growth rates. (f) The error rate of the algorithm as the value of the growth rates $r_1$ and $r_2$ are varied in a fixed ratio $r_2/r_1 = 2$ for one of the systems compared. The horizontal axis shows the value of $r_1$. The other system is held fixed at $r_1 = 2, r_2 = 4$.}

\end{figure}

\section{Assessing performance}

\subsection{General considerations}
The probability that the algorithm correctly detects a difference in measured symmetries depends upon the details of the dynamics underlying the two systems, the number of initial conditions sampled (i.e., the number of transformations applied), the time over which the systems were observed, and the amount of measurement noise present. All of these factors interact with one another, and there is no tractable way to represent, let alone compute, the performance of the algorithm in all conceivable circumstances. Instead, we undertook experiments with simulated data for a couple of nonlinear systems with two aims in mind: (i) to demonstrate that the algorithm does in fact correctly identify whether systems belong to the same dynamical kinds, and (ii) to show how the algorithm behaves as sampling error and the degree of similarity of the underlying dynamics are varied in isolation.

\subsection{Synthetic data}

We tested the algorithm against two classes of simulated system for which we could be certain of the symmetries of the underlying dynamics and thus of whether or not any two systems are of the same dynamical type. The use of simulations also offered control of sampling noise, allowing us to test the robustness of the algorithm against a range of noise distributions. We focused on systems of importance to theoretical ecology, a field in which system identification is notoriously difficult.

\subsubsection{Biological growth}
The first class of systems we considered includes a subset of a theoretically important family of nonlinear models of biological growth, the so-called ``logistic growth models'' \citep{tsoularis_analysis_2002}. Models in this family are described by a differential equation of the form:
\begin{equation}
	\label{eq:genlog}
	\dot{x} = r x^{\alpha} \left(1-(x/K)^{\beta}\right)^{\gamma}
\end{equation}
The variable $x$ refers to the size of a population, while the parameters $r$ and $K$ are generally taken to correspond to an intrinsic growth rate and carrying capacity, respectively. For each value of $\alpha, \beta$ and $\gamma$, Equation \ref{eq:genlog} yields a distinct class of models of growth dynamics. There is significant controversy over which, if any, of these models describes actual biological populations, in part because it is so difficult to distinguish solutions of the different models from one another with standard statistical approaches. We focused on the set of models for which $\alpha = \beta = \gamma = 1$. In this case, Equation \ref{eq:genlog} reduces to the well-known Verhulst logistic equation, $\dot{x} = r x \left(1-(x/K)\right)$. This equation has closed-form solutions which we used to simulate growing populations. Importantly, it is straightforward to solve for the set of dynamical symmetries of population with respect to time. For a transformation, $\sigma$ to be a dynamical symmetry of the Verhulst equation according to Definition \ref{def:DST}, it must satisfy:
\begin{equation}
	\dot{\sigma}(x(t)) = r \sigma(x(t)) \left(1 - \sigma(x(t))/K\right)
\end{equation}

This results in a one-parameter family of symmetries:
\begin{equation}
	\label{eq:LGsym}
	\sigma_p(x) = K x/\left((1-e^{-p}) x + e^{-p} K\right)
\end{equation}
Note that for different values of $K$, a different family of symmetries and thus a distinct dynamical kind is obtained. However, two systems with equal $K$ values are in the same kind no matter how much $r$ differs between them.

\subsubsection{Species competition}

The second class of systems we simulated can be viewed as a generalized logistic growth model that aims to capture the interaction of two or more species competing for resources. Specifically, we looked at the two-species competitive Lotka-Volterra model \citep{pastor_mathematical_2011}:
\begin{equation}
	\label{eq:LV2D}
	\begin{split}
		\dot{x}_1 &= r_1 x_1 \left(1 - (x_1 + \alpha_{12} x_2)/K_1\right)\\
		\dot{x}_2 &= r_2 x_2 \left(1 - (x_2 + \alpha_{21} x_1)/K_2\right)
	\end{split}
\end{equation}

In this case, it's much more complicated to find closed form expressions for symmetries, in part because the symmetries are vector functions, $\pmb{\sigma}(\mathbf{x}) = \langle \sigma_1(x_1,x_2), \sigma_2(x_1, x_2) \rangle$. However, it is straightforward to show that the symmetry condition implies that the symmetries of the system are functions only of the ratio of the growth rates, $r_2/r_1$. We thus tested whether the algorithm correctly classifies systems with the same growth rate ratio in the same dynamical kind despite widely different values of $r_1$ and $r_2$. Conversely, we examined the efficiency of the algorithm at discriminating systems with distinct growth rate ratios despite similar values for all parameters.

\subsection{Classification}

To verify that the algorithm does in fact correctly determine whether systems belong to the same dynamical kind and to explore its limitations in doing so, we performed experiments with both logistic growth and Lotka-Volterra simulations. For growth models, we compared systems governed by the Verhulst equation ($\alpha = \beta = \gamma = 1$) to systems governed by different members of the family of logistic growth models. Specifically, we examined the accuracy of the algorithm for systems with $\alpha = \gamma = 1$ and different values of $\beta$. For each of 10 values of $\beta \neq 1$, we ran 100 trials comparing a system with this value against one in which $\beta = 1$. In all cases, the systems belonged to different dynamical kinds. Of course, the point of the algorithm is not merely to recognize when dynamical systems are different, but to group systems that, despite appearances, belong to the same dynamical kind. So in a second set of experiments, we examined comparisons between two systems governed by the Verhulst equation ($\alpha = \beta = \gamma = 1$) that differed to varying degrees in their respect growth rates, $r$. One system was fixed at $r = 1.5$. We ran 100 trials with the second system at each of 10 different values of $r$. According to Equation \ref{eq:LGsym}, all of these systems belong to the same dynamical kind. In all of these experiments, we imposed measurement noise that was normally distributed with a standard deviation of 5. The results of these experiments are displayed in Figure 2. We conducted an analogous battery of tests with simulated Lotka-Volterra systems. These results are shown in Figure 3.

The principal lesson to be drawn is that the algorithm works -- almost every time it declares two systems to belong to different kinds, they in fact do. Of course, as the error rate curves demonstrate, the more similar the trajectories produced by the dynamics underlying two systems are -- such as when $\beta = 1$ in one system and $\beta \approx 1$ in the other -- the more likely the algorithm is to miss the difference. On the other hand, it's very seldom fooled into declaring distinct kinds when two systems -- no matter how divergent their apparent behavior -- really do belong to the same kind.

\subsection{Noise and normality}

We characterized the robustness of the comparison algorithm under increasing noise in the single-variable case by testing it against pairs of samples from two kinds of Verhulst logarithmic growth system: System 1 for which $K=100$ and System 2 for which $K=90$ (these are in different dynamical kinds since the family of symmetries is a function of $K$). For each of the four possible pairings of System 1 with System 2, we conducted 25 experiments (for a total of 50 experiments in which the systems belonged to the same dynamical kind, and 50 in which they belonged in different kinds). In each experiment, Gaussian noise was added to measurements of the state of each simulated system. This was repeated for 10 different values of the standard deviation, $\sigma$, of the distribution of added noise. A positive example is one in which the two systems actually differ, a negative is when systems of the same type are paired together. Only 8 out of the 100 trials returned a false positive; nearly all errors (by design) were false negatives. The accuracy of the classifier is plotted in Figure 4 as a function of the standard deviation of the noise. The algorithm is $>80\%$ accurate until the standard deviation of the noise reaches more than 12\% of the overall range of the dynamical variable measured.  

We conducted analogous experiments in the multi-variable case by testing pairs of samples from two kinds of Lotka-Volterra system, specifically one for which $r_1=2, r_2=4$ and one for which $r_1 = r_2 = 3$. Again, we ran 100 trials of which 50 involved systems of the same dynamical kind and 50 involved systems of different kinds. We saw comparable or better performance with respect to the preceding experiment (see Figure 4 (c)). There is, however, a subtle property of the algorithm's behavior in this case that bears mentioning. At very low values of noise, it performs no better than at high values of noise. The reason for this is that for two or more variables, there is no precise overlap between the initial, untransformed trajectories of two systems that aren't exactly alike in parameter values. So when the noise is low enough, the algorithm can always fit a joint model essentially by tying together two separate models of functions with disjoint support. However, a little bit of noise blurs the trajectories enough to produce overlap, and the statistics of the joint versus separate fits become dominated by the shapes of the curves, rather than their exact location in phase space. Fortunately, the conservative nature of the algorithm allows this deficiency to be corrected by injecting noise into a sample and watching to see if the algorithm changes its mind from ``same kind'' to ``different''.

Finally, to get a sense for the fragility of the algorithm under violations of normality in the sample noise, we repeated the same experiments, except that a Gaussian noise source was replaced with the skew normal distribution \citep{azzalini_class_1985}. If $\phi(x) = 1/\sqrt{2 \omega^2 \pi} e^{-\frac{(x-\mu)^2}{2 \omega^2}}$ is the standard normal probability density function with mean $\mu$ and standard deviation $\omega$, $\Phi(x) = \int_{-\infty}^x \phi(x')dx'$ is it's corresponding cumulative distribution function, and $T(h, \alpha)$ is Owen's $T$-function \citep{owen_tables_1956}, then the cumulative distribution function of the skew normal is given by $\Psi(x) = \Phi(\frac{(x-\mu)}{\omega}) - 2 T(\frac{(x-\mu)}{\omega}, \alpha)$.  We used  the method of transformation to sample from this skew normal distribution. When the shape parameter $\alpha = 0$, the skew normal distribution reduces to an ordinary Gaussian with mean $\mu$ and standard deviation, $\omega$. We tested the algorithm in experiments in which $\mu = 0$ and $\alpha$ was varied from 0 to 30 while keeping the standard deviation of the distribution fixed at 15 by varying $\omega$ appropriately. The noise was thus increasingly right-skewed. Over this range, however, the skew had no impact on the accuracy of the classifier, as can be seen in Figure 4 (b) and (d). 

\subsection{Dependence on tunable parameters}
\begin{figure}[ht!]
	\label{fig:NS}
	\centering
	\includegraphics[width=\columnwidth]{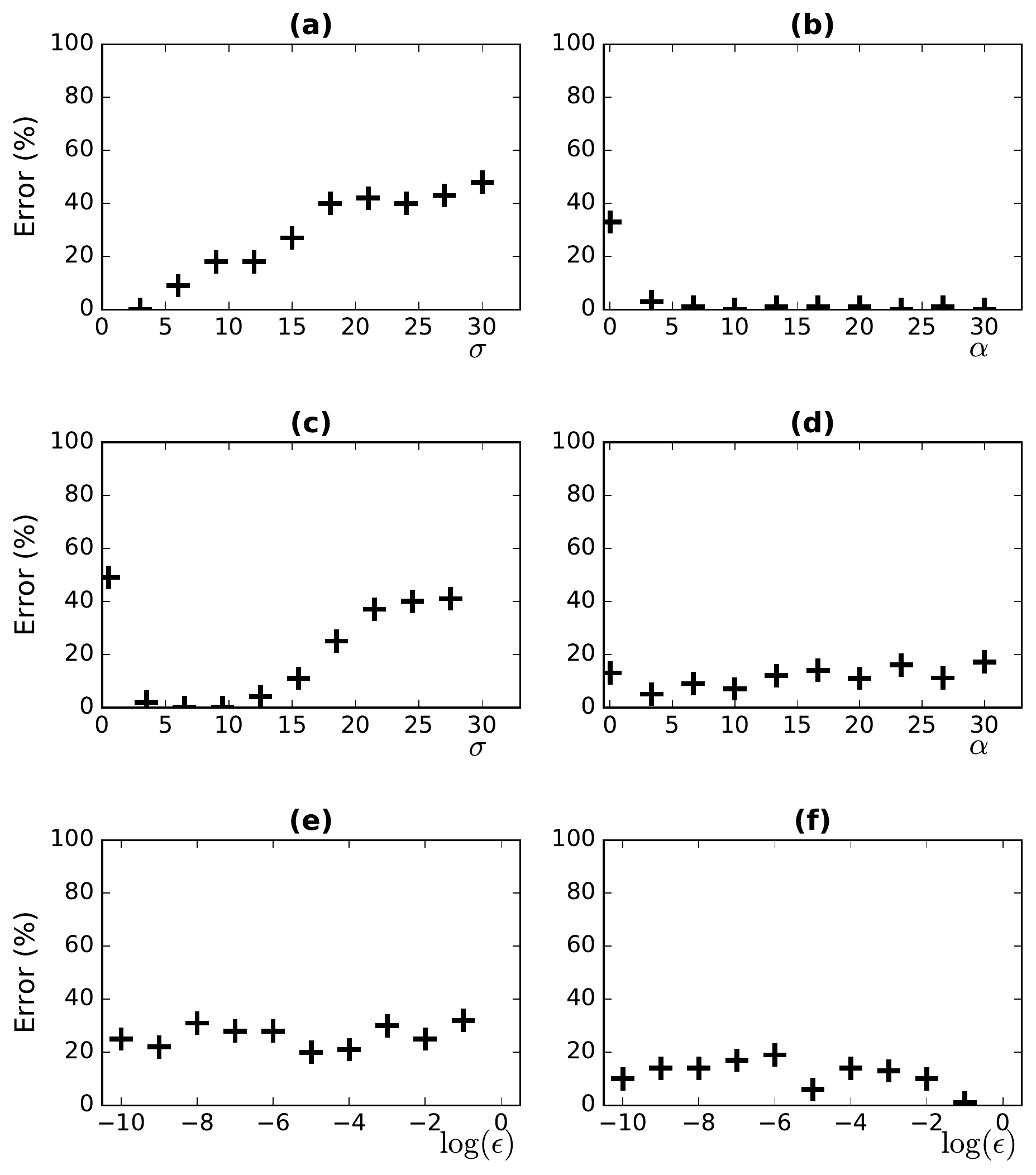}
	\caption{(a) Error rate as a function of standard deviation of normally distributed noise for logistic growth models. (b) Error rate as a function of the $\alpha$-parameter of the skew normal distribution for logistic growth systems. (c) Error rate versus standard deviation of normally distributed noise for two-species Lotka-Volterra systems. (d) Error rate versus $\alpha$ for Lotka-Volterra systems. (e) Error rate as a function of $\log(\epsilon)$ for logistic growth systems. (f) Error rate as a function of $\log(\epsilon)$ for two Lotka-Volterra systems.}
\end{figure}

As indicated in Section \ref{sec:algo} we also examined the sensitivity of the algorithm to the tunable parameter, $\epsilon$, used in a cross-validation routine to choose the order of the polynomial fit for each sampled symmetry. To examine the effect of the value of $\epsilon$ on the classifier error rate, we tested pairs of samples from two kinds of Verhulst logarithmic growth system: System 1 for which $K=100$ and System 2 for which $K=90$ (these are in different dynamical kinds). For each of the four possible pairings of System 1 with System 2, we conducted 25 experiments (for a total of 50 experiments in which the systems belonged to the same dynamical kind, and 50 in which they belonged in different kinds). These experiments were repeated for each of 10 different values of $\epsilon$, spanning 10 orders of magnitude. As is clear from Figure \ref{fig:NS} (e), there is no significant dependence of the overall error rate on the value chosen. We conducted analogous experiments in the multi-variable case by testing pairs of samples from two kinds of Lotka-Volterra system, specifically one for which $r_1=2, r_2=4$ and one for which $r_1 = r_2 = 3$. Again, we ran 100 trials of which 50 involved systems of the same dynamical kind and 50 involved systems of different kinds. Again, there was no significant dependence of the overall error rate upon the value of $\epsilon$, as indicated in Figure \ref{fig:NS} (f).

\section{Discussion}

The experiments reported here demonstrate that it is possible to recognize dynamical kinds without any knowledge of the underlying dynamics, and to do so in the face of significant noise. By directly assessing whether systems share a common set of dynamical symmetries, it is possible to pool systems for model identification or to validate dynamical models without explicitly fitting models to the data. Perhaps most importantly, we have demonstrated that an important sort of scientific inference widely neglected in the literature on automated discovery is amenable to algorithmic treatment. 

But the work reported here only begins to exploit the framework of dynamical kinds. The algorithm described above was designed with a particular class of system in mind, namely those for which the deterministic causal relations are well-described by a differential equation in time. In other words, it was assumed that systems of unknown dynamical type are governed by a deterministic, time-dependent dynamics, and that all random variation observed is noise in the measurement process. But the notion of a dynamical kind as reflected in Definition \ref{def:DST} does not limit one to considering temporal dynamics. And with only modest modification, one can accommodate the general case of stochastic dynamics, that is, causal relations that are inherently noisy, regardless of the measurement process.  The requisite modification is reflected by the following definition:

\begin{definition}[Dynamical symmetry]
	Let $V$ be a set of variables. Let $\sigma$ be an intervention on the variables in $Int \subset V$. The transformation $\sigma$ is a dynamical symmetry with respect to some index variable $X \in V-Int$ if and only if $\sigma$ has the following property: for all $x_i$ and $x_f$, the final probability distribution over $V$ is the same whether $\sigma$ is applied when $E[X] = x_i$ and then an intervention on $X$ makes it such that $E[X] = x_f$, or the intervention on $X$ is applied first, changing its expected value from $x_i$ to $x_f$, and then $\sigma$ is applied.
\end{definition}

To see how this extended definition can be applied in the GCM framework, consider the class of two-variable models in which $x \rightarrow y$, $x$ is distributed according to $p_x(x)$, and the value of $y$ is determined as a function $f$ of the value of $x$ and an additive noise variable:

\begin{equation}
	\label{model}
	y := f(x; y_0) + \eta
\end{equation}

The noise variable, $\eta$ is distributed according to $p_{\eta}(\eta)$ with a mean value of 0. The function $f$ may be any function from values of $x$ to values of $y$ (not necessarily linear). The parameter $y_0$ is the expected value of $y$ when the expected value of $x$ is $x_0$. It's a way of capturing the `memory' and dependence on initial conditions reflected in a differential equation without unfolding such dependence as an infinite chain of discrete time slices. Note that when $f$ is a constant function of $y_0$, then Equation \ref{model} is an instance of the sort of (possibly nonlinear) model considered by \citet{hoyer_nonlinear_2009}.

According to this model, we can express the joint distribution over $x$ and $y$ as:

\begin{equation}
	p(x, y) = p_{x}(x)p_{\eta}(y|x) = p_{x}(x) p_{\eta}(y - f(x; y_0))
\end{equation}

Let's consider $x$ as the index variable and $y$ the target. Then according to the extended definition, a transformation $\sigma(y)$ is a dynamical symmetry just if the joint distribution, $p(x,y)$ is the same whether we (i) first apply $\sigma$ when the expected value $E[x] = x_0$ and then intervene to bring the expected value of $x$ to $E[x] = x_0 + \delta$ (changing the probability density over $x$ to $\tilde{p}_x(x)$); or (ii) bring the expected value of $x$ to $E[x] = x_0 + \delta$ and then apply $\sigma$ to $y$. Given the model expressed by Equation \ref{model}, this condition is satisfied only if:

\begin{equation}
	\begin{split}
		\tilde{p}_x(x)&p_{\eta}(y - f(x; \sigma(y_0))) = \\
	&\tilde{p}_x(x)p_{\eta}(y - \sigma(f(x; y_0)))
	\end{split}
\end{equation}

If we assume that $\tilde{p}_x(x) > 0$, this entails that 
\begin{equation}
	\label{symcon}
	f(x; \sigma(y_0)) = \sigma(f(x; y_0))
\end{equation}

This is exactly the same condition that would have obtained without the additive noise in the model. In general, deterministic systems governed by first-order differential equations can be cast in the form of Equation \ref{model} without the additive noise (and it's not difficult to generalize that form for higher-order dynamics). By adding the noise term, we can consider stochastic versions of the dynamical systems considered above. For example, we can introduce a stochastic logistic growth model this way:

\begin{equation}
	\label{stoclog}
	y:=\frac{(K-x_0)y_0 x}{(K-y_0)x_0 + (y_0-x_0) x} + \eta
\end{equation}
The variable $y$ represents the population at $t + \delta$ which is dependent upon $x$, the population at $t$. The variable $y_0$ is the population at $t + \delta$ that results from a population of $x_0$ at $t$. Though it is perhaps not obvious from the form, note that Equation \ref{stoclog} reduces to the deterministic model when the noise term is set to 0. The symmetry condition expressed by Equation \ref{symcon} entails in this case that 
\begin{equation*}
	\begin{split}
	\sigma &\left(\frac{(K-x_0)y_0 x}{(K-y_0)x_0 + (y_0-x_0) x}\right) =\\
	& \quad \quad \frac{(K-x_0)\sigma(y_0) x}{(K-\sigma(y_0))x_0 + (\sigma(y_0)-x_0) x}
\end{split}
\end{equation*}
which is in turn satisfied by
\begin{equation*}
	\sigma_p(y) = \frac{K y}{(1 - e^{-p}) y + e^{-p} K}
\end{equation*}
This is exactly the same set of symmetries that characterize the deterministic dynamical kind (see Equation \ref{eq:LGsym}). In other words, the extended definition of dynamical symmetry picks out an expanded class of systems that includes as a proper subset the systems governed by a deterministic Verhulst equation with a particular value of the carrying capacity, $K$. But it also includes systems with stochastic causal relations as described by Equation \ref{stoclog}. 

Despite this overlap in conditions for kind membership, detecting sameness or difference of dynamical kind in systems exhibiting stochastic causation requires an approach different from the algorithm described above. That's because individual trajectories are random walks biased by the function $f$ rather than noisy instantiations of deterministic trajectories. We leave the development of methods appropriate for stochastic causation to future work. But we would like to reiterate the generality of the theory that has allowed us to frame the problem of kind discovery in this extended context. We have already demonstrated an algorithm for the causal systems of the sort that are the focus of most work in system identification. The point we wish to stress is that the theoretical framework of dynamical kinds extends to the relations of stochastic causation at the focus of work in causal discovery. Exploiting the framework in this context in order to infer scientifically salient kinds, validate models, and pool data for model construction will require only modest algorithmic innovation.

\subsubsection*{Acknowledgements}
This material is based upon work supported by the National Science Foundation under Grant No. 1454190.

\bibliographystyle{chicago}
\bibliography{UAIpaper}

\end{document}